\documentclass{article}

\usepackage{amsmath}

\usepackage{arxiv}
\usepackage[utf8]{inputenc} 
\usepackage[T1]{fontenc}    
\usepackage{hyperref}       
\usepackage{url}            
\usepackage{booktabs}       
\usepackage{amsfonts}       
\usepackage{nicefrac}       
\usepackage{microtype}      
\usepackage{lipsum}
\usepackage{graphicx}
\graphicspath{ {./images/} }
\usepackage{algorithm}
\usepackage{algpseudocode}

\usepackage{adjustbox}
\usepackage{multirow}
\usepackage{multicol}
\usepackage{amssymb}
\usepackage{bbm}

\usepackage{tikz}
\usetikzlibrary{shapes,shapes.geometric,matrix}
\usepackage{anyfontsize}
\usetikzlibrary{arrows.meta}
\usetikzlibrary{shapes.symbols}
  
\title{Multiple Inputs Neural Networks for Medicare fraud Detection
}

\author{
  Mansour Zoubeirou A Mayaki \\
  Université Côte d’Azur \\
  CNRS, Inria, I3S \\
  France\\
   \And
  Michel Riveill \\
  Université Côte d’Azur \\
  CNRS, Inria, I3S \\
  France\\
}

\begin{document}
\maketitle

\begin{abstract}
Medicare fraud results in considerable losses for governments and insurance companies and results in higher premiums from clients. Medicare fraud costs around 13 billion euros in Europe and between  21 billion and  71 billion US dollars per year in the United States. This study aims to use artificial neural network based classifiers to predict medicare fraud. The main difficulty using machine learning techniques in fraud detection or more generally anomaly detection is that the data sets are highly imbalanced. To detect medicare frauds, we propose a multiple inputs deep neural network based classifier with a Long-short Term Memory (LSTM) autoencoder component. This architecture makes it possible to take into account many sources of data without mixing them and makes the classification task easier for the final model.
The latent features extracted from the LSTM autoencoder have a strong discriminating power and separate the providers into homogeneous clusters.
We use the data sets from the Centers for Medicaid and Medicare Services (CMS) of the US federal government. The CMS provides publicly available data that brings together all of the cost price requests sent by American hospitals to medicare companies.
Our results show that although baseline artificial neural network give good performances, they are outperformed by our multiple inputs neural networks. 
We have shown that using a LSTM autoencoder to embed the provider behavior gives better results and makes the classifiers more robust to class imbalance. 

\end{abstract}


\section{Introduction}
The progress made in the field of big data
and data management makes it possible to fight fraud more effectively in several business sectors such as finance, banking and insurance. Insurance fraud results in considerable losses for governments and insurance companies and results in higher premiums from clients. According to Insurance Europe, detected and undetected fraud would cost European customers and insurers around 13 billion euros \cite{insueuro} per year.
In the field of medicare, in France the compulsory scheme detected 261.2 million euros of fraudulent services in 2018, mainly due to providers (opticians, pharmacists, medical auxiliaries, doctors, etc.) and institutions \cite{bilan2018}. In the United States, according to the Federal Bureau of Investigation (FBI), fraud represents $ 5-10\% $ of medicare claims and costs insurance companies between $ 21$ billion and $ 71$ billion per year \cite{johnson2019medicare}. In a context where reducing management costs is a real issue for health insurers, the fight against fraud is a real expectation so that everyone receives a fair return for their contributions.
Most insurance companies have business rule based fraud detection systems. These methods, although effective, are often very difficult to set up and maintain. Indeed, a rule-based fraud detection system constantly requires the presence of experts in the field and constant updates of the rules.

Models based on statistical methods and machine learning make it possible to automatically build patterns and thus detect fraudulent activities effectively.
In the field of medicare, insurance companies receive requests for reimbursement (claims) sent by healthcare providers on behalf of their clients. For example, when an insured buys new corrective lenses, the optician sends the invoice to his mutual medicare which covers a percentage of the costs (for example 90\%). We find the same circuit in almost all other branches of insurance: auto insurance, business, home etc. The most common types of fraud include billing for appointments that the patient has missed, billing for services that are more complex than those performed, or billing for services not provided. Thus, in medicare, it happens that customers or healthcare providers send false claims to insurance companies in order to benefit from reimbursements. There are several other types of fraud that insurance companies face.

The main difficulty in applying machine learning techniques in fraud detection or more generally anomaly detection is that you don't have enough data labeled as anomalous or fraudulent. Thus, you end up in a situation of  imbalanced class where one class is very poorly represented compared to the others. 
For example in medicare, fraudulent transactions often represent less than $5\%$ of all transactions. 
The high imbalance rate makes it very difficult for machine learning algorithms to learn as they will tend to favor the majority class. 
Another challenge  is that the labels are often not accurate. Some samples labeled fraudulent may not be real frauds and and vice versa.

In this study we use publicly available medicare data sets from the Centers for Medicare and Medicaid Services (CMS) for period 2017–2019 \cite{CMSGov2018}.
The data sets contain the hospitalization requests (Inpatient Data), the outpatient care requests (Outpatient Data) and the claims details. We also use the Office of Inspector General's list of excluded individuals and entities (LEIE) \cite{leie2020}. The LEIE table contains the list of healthcare providers excluded from the healthcare system for illegitimate or fraudulent activities.
 Since we are interested in fraud detection, we call the group of fraudsters the positive class and non-fraudulent group the negative class. The main challenge working with this data set is that it is highly imbalanced with a fraud rate between $0.038 \%$ and $0.074 \%$.
 Another challenge is that it exhibits big data properties. Each year the CMS releases approximately 9 millions records from multiple sources, mixed-type and high-dimensional. 

To deal with class imbalance issue, there are two main approaches with varying performance depending on the field of application and the complexity of the problem: the resampling approach (or data level) which consists in balancing the classes by adding or removing data from classes  and the approach which consists in modifying the learning algorithms so that they take into account the class imbalance (algorithm level). 

To detect medicare frauds, we propose a  multiple inputs deep neural networks based classifier with a Long-short Term Memory (LSTM) autoencoder component. We call this architecture \textbf{MINN-AE}. This architecture makes it possible to take into account many sources of data without mixing them and makes the classification task easier for the final model. The LSTM autoencoder part of MINN-AE plays a dimension reduction role for the provider data and its latent vector describes the provider behavior over time. 
The rest of the paper is outline as follows. The \textbf{Related works} section discusses the other studies and articles related to imbalance data handling, deep learning for anomaly and medicare fraud detection. In section \textbf{Data sets}, we describe the data sets used in this study and the data pre-processing steps. The \textbf{Classifiers} section describes in details our models architecture, the loss functions we tested and the hyperparameters optimization steps. The results are presented and discussed in section \textbf{Results and discussions}.

\section{Related works}

The work presented here does not only concern research in the fields of medicare or fraud detection. We also present some techniques proposed to remedy the problem of class imbalance. These two concepts are inseparable because in fraud detection we always face the problem of class imbalance.

\subsection{Medicare Fraud Detection and Resampling Methods}
The Centers for Medicare and Medicaid Services (CMS) data has been used in numerous studies to detect medicare fraud. Most of these studies use resampling techniques to overcome the imbalance class issue ( Bauder et~al. \cite{bauder2018detection}; Liu et~al. \cite{liu2013healthcare}; Herland et~al. \cite{herland2018big}; Johnson et~al.\cite{johnson2019medicare}; Van et~al. \cite{van2007experimental}). 
In there study, Herland et~al.\cite{herland2018big} show that the combination of the three parts of CMS data makes it possible to detect more precisely fraudulent activities. They compared the performances of logistic regression, random forest and gradient boosting classifiers on each part of the data taken separately with those obtained grouping all the parts and results show that the performance of all classifiers improves dramatically using all parts of the data, and that logistic regression outperforms all other models. Using the CMS data from 2010, Liu et~al. \cite{liu2013healthcare} added some geo-location information to detect fraud. They went from the hypothesis that medicare beneficiaries are senior, disabled or poor and prefer to choose the health service providers locating in a relatively short distance and if the distance between the providers location and the client living place is too long, it may imply a fraud. 
Bauder et~al. \cite{bauder2018detection} used three different classifiers to detect fraudulent medicare provider claims: C4.5 decision tree (C4.5), Support Vector Machine (SVM), and Logistic Regression (LR). They used the CMS data over the period 2012-2015 combined with the Office of Inspector General's list of excluded individuals and entities (LEIE) \cite{leie2020}. The authors also used random undersampling technique to handle the class imbalance problem. Their results show that the C4.5 decision tree and logistic regression classifiers have the best performance on detecting fraud.
In their study, Johnson et~al.\cite{johnson2019medicare} compared six resampling techniques for imbalanced classes  using the CMS data over the period 2012–2016 \cite{CMSGov2018}. These authors combined artificial neural network models with class imbalance techniques to predict fraud. They tested random undersampling (RUS), random oversampling (ROS), mean square error (MSE) and Focal Loss techniques among others.
According to their results, RUS improves the performance of the classification algorithm if the majority class share is above $99\%$. The authors then conclude that maintaining sufficient representation of the majority class is more important than reducing the level of class imbalance, and that down-sampling until classes are balanced can deteriorate classification performance.
Van et~al. \cite{van2007experimental}, in another study also compared different resampling techniques using 11 types of classifiers. In their experiments, they used 35 different data sets with degrees of imbalance (ratio between the number of samples in the minority class and that of the majority class) varying between $1.33\%$ and $35 \%$.
The resampling techniques used in this article are: random undersampling (RUS), random oversampling (ROS), one-sided selection (OSS), cluster-based oversampling (CBOS), Wilson's editing (WE), synthetic minority oversampling technic (SMOTE) and borderline-SMOTE (BSM). To facilitate comparison of the results, they grouped the data sets into 3 categories according to the percentage of data in the minority class: $ <5 \% $, between 5 \% and 10 \%, $> 10 \% $. Their results show that whatever the category considered, the RUS technique tends to give better performances ($32\%$ of the time).

Most of these studies come to the conclusion that undersampling (down-sampling) is more efficient than over-sampling. These results go against what one might have expected as undersampling often leads to a loss of information. One possible explanation is that in some situations, adding new artificial data will add more noise than useful information to the model. Depending on the complexity of the problem (or data), it is necessary to test the two approaches (down-sampling and over-sampling) to see which one fits best.

\subsection{Algorithm Level Methods for Imbalanced Classes}
To overcome the problem of class imbalance, some authors propose to alter the learning algorithm in the way that it takes into account the problem (Wang et~al. \cite{wang2016training}; Haishuai et~al.\cite{wang2018predicting}; Lin et~al.\cite{lin2017focal}) . The main idea of algorithm level method is to modify the learning algorithms so that they give more importance to the samples from the minority class which is often the class of interest.

Lin et~al.\cite{lin2017focal} proposed an algorithm level method which consists in rewriting the classical entropy loss function by integrating two new parameters: $\alpha$ takes into account the imbalanced issue and $\gamma$ (gamma) the complexity of classifying the samples. This new loss function called \textbf{Focal Loss} is obtained by multiplying the classical cross entropy (CE) by a modulation factor $\alpha(1-p)^{\gamma}$. hyperparameter $\gamma \geq 0$ adjusts the rate at which easy examples  are  down  weighted and  $\alpha$  is  a  class-wise  weight used to give more importance to the minority class \cite{johnson2019survey}. Lin et~al. \cite{lin2017focal} applied their new cost function (Focal Loss) to object detection in images and their results show that this loss function gives better performance than most benchmark models. Wang et~al. \cite{wang2016training} proposed another algorithm level method called \textbf{mean false error} (MFE) which consist in decomposing the classical mean squared error (MSE) in two components in other to give more weights to the minority class samples. 
They rewrite the classical MSE as a kind of weighted average of the errors of the two classes. In this way, all the classes participate equally in the final loss function.
Haishuai et~al.\cite{wang2018predicting} in their paper used an artificial neural network based model with a \textbf{cost matrix} to predict readmission of patients from a hospital. 
They defined a \textbf{cost matrix} such that the cost of misclassified readmission (False negative) is greater than that of misclassified non-readmission (false positive). This technique can be seen as an algorithm level method because during optimization, the model will tend to penalize more or give more weight to the minority class (readmission) samples in the loss function. 

Algorithm level methods often give better results than data level methods as they don't alter the training data and don't lead to a loss of information. However in some situations, when you don't have enough data, oversampling can be a good way to extend your data set. Moreover, when the distribution of the samples in the majority class is stationary (the samples are very close to each other) undersampling may work very well as we don't loose lot of information by deleting some samples.

\section{Data sets}
\label{data}
The Centers for Medicare and Medicaid Services (CMS) \cite{CMSGov2018} is an important source of data for research on fraud detection in medicare. The CMS publishes a series of publicly available data each year containing information on the use and payments of medical procedures, services and prescription drugs provided to beneficiaries as well as data on physicians and other actors in the healthcare system. These CMS data combined with the Office of Inspector General's list of excluded individuals and entities (LEIE) \cite{leie2020} containing the list of healthcare providers excluded from the healthcare system for illegal activity allows researchers in the field of statistics and artificial intelligence to propose new methods to fight against fraud in medicare.
One of the main challenges when working with these data sets is that they exhibit big data properties. Each year the CMS releases approximately 9 millions records from multiple sources, mixed-type features and high-dimensional. 
In this study we use publicly available medicare data sets from the Centers for Medicare and Medicaid Services (CMS)  for the period 2017–2019. The dataset can be downloaded from the CMS website \cite{CMSGov2018}.

The CMS data sets contains mainly three types of information: hospitalization requests (Inpatient Data), outpatient care requests (Outpatient Data) and beneficiary information (Beneficiary Details Data). 
The Inpatient Data table contains information on patients admitted to hospitals. Table Outpatient Data gathers information on patients who have visited the hospital without being hospitalized there.
These tables have been combined into a single table containing patient information, provider information and claims details (see Fig. \ref{process}). We thus have information such as the identifier of the beneficiary (BeneID), the identifier of the claims (ClaimID), the unique identifier of the healthcare provider (NPI), the refunded amount (AmtReimbursed) etc. Records  within the  data  set also  contain  various  provider-level  attributes,  e.g.  National  Provider  Identifier  (NPI), first and last name, gender, credentials, address etc. Fig.~\ref{process} shows the aggregating process used. 

\tikzset{
  startstop/.style = {rectangle, rounded corners, minimum width=5cm, minimum height=1cm,text centered, draw=black, fill=white,line width=1pt}
,
box/.style = {rectangle, rounded corners, minimum width=2cm, minimum height=2cm,text centered, draw=black, fill=gray!10,line width=2pt},
block/.style={rectangle, draw=black, thick, minimum width=2cm, fill=white,text centered,text width=10em,  minimum height=5cm,line width=1pt},
block2/.style={rectangle, draw=black, thick, minimum width=80cm, fill=white, minimum height=60cm,line width=1pt}
}
            
\begin{figure}[htbp]
\begin{center}
\resizebox{0.5\columnwidth}{!}{%
\begin{tikzpicture}[node distance=2cm , auto,minimum size=2cm]

\node [block,scale=1] (table1) {\large \textbf{Provider detail} \\
\textcolor{red}{Provider ID (NPI}\\
First name\\
Last name\\
Gender\\
Credentials\\
Address etc.}; 
\node [block,scale=1, right of=table1, xshift=3cm] (table2) {\large\textbf{Inpatient datal} \\
\textcolor{red}{Claim ID}\\
\textcolor{red}{Beneficiary ID}\\
Provider ID\\
Admission date\\
Discharge date\\
Attending physician\\
Operating physician\\
Reimbursed amount etc.};          
\node [block,scale=1, right of=table2, xshift=4cm] (table3) {\large\textbf{Outpatient datal} \\
\textcolor{red}{Claim ID}\\
\textcolor{red}{Beneficiary ID}\\
Provider ID\\
Admission date\\
Discharge date\\
Attending physician\\
Operating physician\\
Reimbursed amount etc.};    

\node [startstop,scale=1, below of=table2,xshift=3cm, yshift=-4cm] (merge1) {Merge on common columns};         
\node [startstop,scale=1, below of=merge1,xshift=0cm, yshift=-2cm] (merge2) {Merged Inpatient and Outpatient data};    
\node [startstop,scale=1, left of=merge2,xshift=-6cm, yshift=0cm] (merge3) {Merge on Provider ID};  
\node (db1) [cylinder, 
        fill=blue!30,shape border rotate=90, 
        draw,minimum height=2cm,
        minimum width=2cm,
        shape aspect=.25, left of=merge3,xshift=-2cm, yshift=3cm] {\large Final table};
        
\draw [->, draw=black] (table2) |- (merge1);
\draw [->, draw=black] (table3) |- (merge1);
\draw [->, draw=black] (merge1) -- (merge2);
\draw [->, draw=black] (merge2) -- (merge3);
\draw [->, draw=black] (merge3) -| (db1);
\draw [->, draw=black] (table1) -- (merge3);

\end{tikzpicture}
}
\end{center}
\caption{Data merging flow. The tables are aggregated using unique identifiers highlighted in red. }
\label{process}
\end{figure}
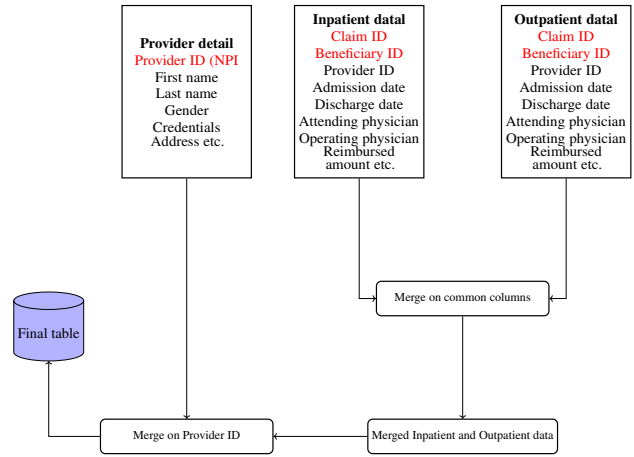


Note that for a fraudster provider, we do not know which of its claims are fraudulent and which are legitimate. To overcome this label issue, we considered that if a provider is fraudster, all its claims are also fraudulent \cite{johnson2019medicare}. This assumption makes sense because if a provider has been declared as a fraudster, the decision certainly comes from a deep analysis of his recent activity and his claims reflect illegal activities.
 We created additional features for the providers by aggregating the variables at the invoice level. For each provider we created new variables by taking the mean, the variance, the sum, the skewness coefficient of the numerical variables per trimester.
In order to capture the provider behavior over time, we use a slicing technique called bucketing \cite{wang2018predicting}. The behavior of each provider with respect to each variable can be considered as a time series. Indeed, over the years the provider makes several claims for  different patients. For example, we can calculate the total amount paid by the insurance company to the provider each month. The bucketing technique consists of separating the claims from each provider into groups according to time and aggregation indicators are calculated in each group. There are two possible levels of aggregation: first order (first order features) and second order (second order features). The first order indicators are mean, standard deviation, variance, sum, maximum, minimum, skewness and kurtosis. The second order indicators are: Energy  (E1),  Entropy  (E2),  Correlation  $(\rho_{x,y})$,  Inertia  and Local Homogeneity (LH). In this study, we only calculate the first order indicators for each numeric variable per trimester. The data of each provider over each year is separated into 4 blocks. In each block, the aggregation variables are calculated: mean, standard deviation, skewness, maximum, minimum, sum etc. After cleaning and preprocessing, the final data set has a  fraud rate of $0.5\% $. 
Table \ref{tab:ps_data_us} shows a subset of the provider level data and Table \ref{leie} a subset of the LEIE data.

\begin{table}[htbp]
 \caption{Subset of the agregated data set on provider level}
 \label{tab:ps_data_us}
\begin{center}
\resizebox{0.7\columnwidth}{!}{%
\begin{tabular}{|l|l|l|l|l|l|}
\hline
 NPI &  BeneID count &  DeductibleAmtPaid  mean &  InscClaimAmtReimbursed  sum &  Fraud \\
\hline
 1000051001 &   24 &  213.60 & 104640 &  No \\
 1000051101 &  117 & 502.16 &  605670 &     Yes \\
 1000051005 &  138 &   2.08 &   52170 &  No \\
 1000051007 &  58 & 45.33 &  33710 &    No \\
 1000051009 &  36 &  53.86 &  35630 &   No \\
\hline
\end{tabular}}
\end{center}
\end{table}

\begin{table}[htbp]
\caption{Subsset of LEIE data set}
\label{leie}
\begin{center}
\resizebox{0.7\columnwidth}{!}{%
\begin{tabular}{|l|l|l|l|l|}
\hline
NPI &          CITY & STATE &  EXCLTYPE &  EXCLDATE \\
\hline
 1306111111 &   GARDEN CITY &    NY  &   1128a1 &  20180220 \\
 1306111111 &        WARREN &    OH  &   1128b5 &  20190220 \\
 1306111111 &  PHILADELPHIA &    PA  &   1128b7 &  20191231 \\
 1306111111 &      FLUSHING &    NY &   1128a1 &  20190220 \\
 1306111111 &   SPRINGFIELD &    MO &   1128b4 &  20200220 \\
\hline
\end{tabular}
}
\end{center}
\end{table}

Ultimately, we have three sources of data to predict fraud: the first table which contains all information on the claims (total amount, equipment, amount of each equipment etc.), the second table contains aggregated data per year for each provider (total amount on the year, number of each type of equipment, number of patients, etc.) and the third table contains aggregated data per trimester (total amount per trimester, number of clients per trimester etc.) for each provider.

\section{Methodology }
In this section, we present our \textbf{MINN-AE} model and the other classifiers we tested. We compared MINN-AE to baseline artificial neural networks and non neural networks models.
\label{classifiers}

\subsection{State-of-the-art Classifiers}
We compared the artificial neural network models to three state-of-the-art classifiers : logistic regression, random forest and gradient boosting. These three classifiers are good baseline models for classification tasks.
The \textbf{logistic regression}  makes it possible to predict the probability of an event happening (fraud value of 1) or not (non fraud value of 0) from the optimization of the regression coefficients. This result always varies between 0 and 1. When the predicted value is greater than a threshold, the event is likely to occur, while when this value is below the same threshold, it is not. \textbf{Random forest} makes an aggregation of several decision trees classifiers trained on slightly different data subsets. It thus makes it possible to have results that are robust to variations and generally gives good performance in inference.
\textbf{Gradient boosting} is another method of aggregating decision trees. Decision trees are built sequentially by giving more and more weight to misclassified data \cite{friedman2002stochastic}. The algorithm thus uses the errors of the present model to fit the future model.
We also tested another implementation of Gradient boosting algorithm called \textbf{XGBoost }. The XGBoost implementation is computational efficiency and often better model performance. In this study, the hyper parameters were chosen using a grid search.

We chose these three classifiers because they are commonly used and provide reasonably good performance. We compare their performance to those of of artificial neural networks based classifiers. 
The selected hyper parameters are listed in the following table \ref{classics}:
\begin{table}[!h]
\caption{State-of-the-art classifiers hyper parameters}
\label{classics}
\begin{center}
\resizebox{\columnwidth}{!}{%
\begin{tabular}{|l|c| c|c|c|c|}
 \hline
 {Models} &  n\_estimators &  min\_samples\_split &  max\_features &  max\_depth & min\_samples\_leaf \\
  \hline
     Random forests  & 50 & 20 &- & 15 &- \\
 \hline
 Gradient Boosting  & 80 & 20 & auto & 10 & 20\\

\hline
\end{tabular}}

\end{center}
\end{table}
\subsection{Baseline Artificial Neural Network}

We first tested some baseline Multi-Layer Perceptrons (MLP) models consisting of a single input layer, multiple hidden layers, and an output layer.
These models take an invoice as input and predicts if it's fraud or not. The number of layers and the number of neurons in each layer are variables (hyperparameters) that must be chosen carefully for neural network models to give good results. These variables remain constant throughout the training process and have a direct impact on the performance of the models. 
We refer to the baseline neural network as \textbf{BNN}. The BNN is a simple multilayers perceptron model where all the features are concatenated and used as input for the model. We tested some version of BNN using the loss functions as describe in subsection \ref{losses}. \textbf{BNN weighted} stands for BNN with weighted loss, \textbf{BNN focal} with focal loss, \textbf{BNN mfe} with the mean false error loss and $\textbf{BNN rus}$ the best BNN obtained by random under sampling. 
The choice of hyperparameters is described in the next subsection \ref{hyper_params}.

\subsection{Our \textbf{MINN-AE} Model's Architecture}
\label{model_multi}
\textbf{MINN-AE} is made up of two different inputs layers. The first input layer receives the data relating to the claims and the second input layer receives the data relating to the healthcare provider. The model is thus composed of two blocks which meet at the end. Each block consists of an input layer, hidden layers and an output layer. The outputs of the two blocks are then concatenated to form a single vector. Such an architecture makes it possible to simultaneously take into account data on claims and those on healthcare provider separately.
In our version of the multi-input model, the second block is a Long-short Term Memory (LSTM) autoencoder. We first trained the LSTM autoencoder on the provider level data. This autoencoder learns to reconstitute a healthcare provider behavior over time. Then we used the latent vector from the LSTM autoencoder as an input vector for our final model. The final model is thus composed of an input layer which takes as input the claims details and another input layer which makes it possible to inject the latent vectors coming from the autoencoder. In this architecture, the autoencoder plays a dimension reduction role for the provider data and its latent vector describes the healthcare provider behavior. Note that the autoencoder parameters remain constant when learning the final model. 
The model's architecture is presented in Fig.~\ref{MINN} below.

	

\begin{figure*}[h!]
\begin{center}
\centerline{\includegraphics[width=\textwidth]{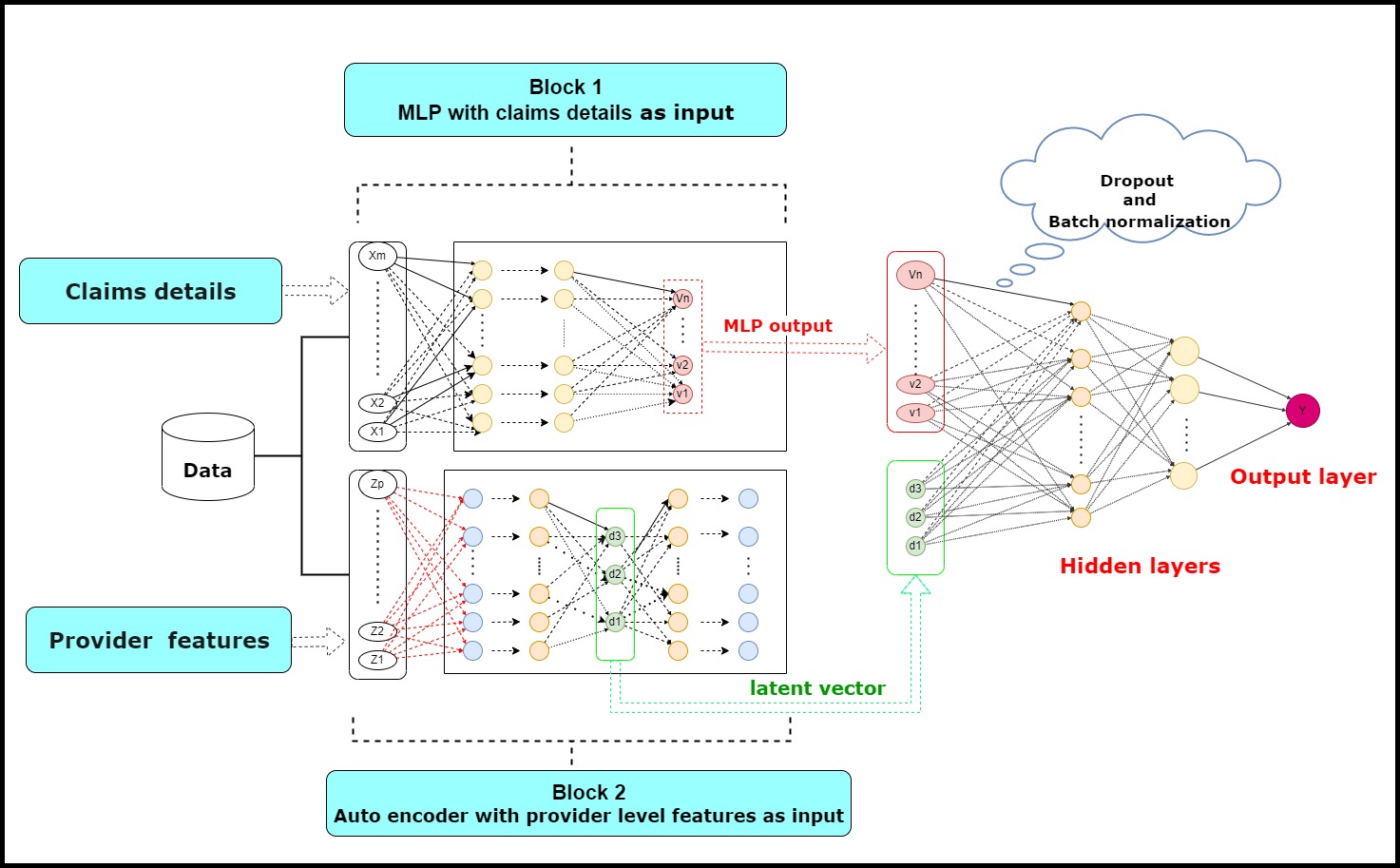}}
	\caption{Visualization of the proposed neural network architecture. Block 1 receives features related to the invoice. Block 2 receives features related to the provider behavior and trains a LSTM based autoencoder. The latent vector of the autoencoder and the output of block 1 are concatenated and used as input for the next hidden layers of the model. }
	\label{MINN}
\end{center}	
\end{figure*}
\subsection{Performance Metrics}
The classifiers are evaluated using the precision-recall curve (PRC). This plot shows precision values for corresponding recall values. It provides a model-wide evaluation like the receiver operating characteristic (ROC) plot or the cost curve (CC). The  area under the curve (AUC) score of precision-recall curve, denoted as AUC (PRC), is likewise effective in multiple-classifier comparisons \cite{saito2015precision}. 
The AUC (PRC) measures the entire two-dimensional area under the entire precision-recall curve (by integral calculations) from (0,0) to (1, 1). 
We don't use AUC (ROC) as most studies do because as Saito et~al. \cite{saito2015precision} show in their study, the  AUC (ROC) is not well suited in case of imbalance class.
In their article \cite{saito2015precision}, these authors showed that AUC (ROC) could be misleading when applied in imbalanced classification scenarios instead AUC (PRC) should be used. Their study showed via multiple simulations that AUC (ROC) fails to capture the variation in class distribution contrary to the AUC (PRC).
As the AUC (ROC) is used as performance metric in most  studies in the literature, we will give its value for each of our classifiers just as an indication.
In order to have more detail on classifiers performance, we also compute the precision and the G-means score. The precision gives the performance of the classifier on the positive class and the G-mean metric makes a compromise between the true positive rate TPR or recall and the true negative rate (TNR).
$$TNR=\frac{TN}{TN+FP} \hspace{1cm} G\-mean=\sqrt{Recall \cdot TNR} $$

$$ Recall=\frac{TP}{TP+FN} \hspace{1cm} Precision=\frac{TP}{TP+FP} $$
\subsection{Hyperparameters Optimization}
\label{hyper_params}
We used the version of the mini-batch stochastic gradient descent (SGD) called SGD Adam with a batch size of 256. This optimizer allows to adapt the step of the gradient during the training of the model and is known to give better performances compared to other versions of SGD \cite{johnson2019medicare}. We kept the default values of the other hyperparameters: lr=0.001, $\beta_{1}=0.9$ and $\beta_{1}=0.999$. The rectified linear unit (ReLU) activation function is used in neurons of the hidden layers, and the sigmoid activation function is used for output layer to estimate posterior probabilities.

We choose the best hyperparameters using the \textbf{KerasTuner} library and on a small holdout set ( $10\%$) of the validation dataset. KerasTuner is an easy-to-use, scalable hyperparameter optimization framework that solves the pain points of hyperparameter search \cite{omalley2019kerastuner}. The AUC (PCR) metric is used to compare the set of hyperparameters. 
We also added between the hidden layers a Batch Normalization layer followed by a dropout layer. Batch Normalization \cite{ioffe2015batch}  makes artificial neural network faster and more stable by normalizing and rescaling layers inputs. Dropout consists in "deactivating" randomly some neurons during training \cite{srivastava2014dropout}. Each neuron being possibly inactive during a learning iteration, this forces each unit to "learn well" independently of the others and thus avoid overfitting. The best architecture for each model obtained using Keras tunner are listed in Table \ref{ann_architectures}. 
\begin{table}[htbp]
\caption{Neural network architecture for each data set.}
\label{ann_architectures}
\begin{center}
\begin{tabular}{|c|c|c|c|}
\hline
{\textbf{Data set}} & \textbf{architecture} & \textbf{Hyper-parameters} &  \textbf{Epochs} \\
 \hline
BNN &MLP& (45,25,1) & 50  \\
BNN weighted &MLP& (30,15,1) & 50    \\
BNN focal&MLP& (30,12,1) & 50 \\
BNN MFE   &MLP&   (25,10,1) &  50 \\
BNN RUS  &MLP&  (15,5,1) & 50 \\
\hline

\multicolumn{1}{|l}{\multirow{2}{*}{\textbf{MINN-AE (ours)}}} & MLP &  (20,10,1)+(15,1) &50  \\
\multicolumn{1}{|l}{}& LSTM autoencoder  &  (64,32,10,32,64) &  30   \\
\hline
\end{tabular}
\end{center}
\end{table}

\subsubsection{Optimal Decision Threshold}
The output of an artificial neural network is the posterior probability of a sample being fraudulent. To classify each sample, we need a decision threshold above witch the samples are labeled as fraudulent.  The choice of the optimal decision threshold is made on the validation data during the training phase by cross validation. Thus during each iteration, we choose the optimal threshold by varying it between 0 and 1. We test several values between 0 and 1 and we choose the threshold which maximizes the G-Mean score on the validation data. At the end of the training, we end up with 10 threshold values corresponding to the optimal values of the 10 models trained during $k = 10$ folds (iterations). The 10 values are then averaged to have the final optimal threshold that will be used during inference \cite{johnson2019medicare}.
\subsection{Loss Functions}
\label{losses}
In this subsection, we describe the loss function tested with our classifiers.
\subsubsection{Weighted Cross Entropy}
This cost function integrates class-wise weights. The loss of each data is multiplied by the weight of the class it belongs to. The total cost function is written as follows:

\begin{equation}
    Weighted \\\ CE \\\ loss=-\sum_{i=1}^{C}w_{i}P_{i}\log(P_{i})
\end{equation}
With $w_{i}$ the weight associated to class $i$, $P_{i}$ the probability of class $ i $ and $ C $ the total number of classes. The  class weights are calculated  as follows \cite{tensorflow2015-whitepaper} :

$$
class\_0 = \frac{1}{neg}.\frac{total}{2}
\hspace{0.5cm}
class\_1 = \frac{1}{pos}.\frac{total}{2}
$$
 $total$ refers to the total number of samples, $pos$ numbre of samples in the positive class  et $neg$ number of samples in the negative class. Multiplying by the $\frac{total}{2}$ makes sure that the loss function keeps the same amplitude for each sample. 

\subsubsection{Focal Loss}
To address class imbalance, Lin et~al.\cite{lin2017focal} propose to rewrite the entropy function by integrating two new parameters: $\alpha$ takes into account the imbalanced issue and $\gamma$ (gamma) the complexity complexity of classifying samples. This new loss function called Focal Loss is obtained by multiplying the cross entropy (CE) by a modulation factor $\alpha(1-p)^{\gamma}$. hyperparameter $\gamma \geq 0$ adjusts the rate at which easy examples  are  down  weighted and  $\alpha $  is  a  class-wise  weight used to give more importance to the minority class \cite{johnson2019survey}.
The proposed loss function is written as follows:
$$FL(p)=\alpha(1-p)^{\gamma}\log p$$
For easy classified samples ($ p -> 1 $) the modulation factor tends towards $ 0 $ which reduces their importance in the final loss function. Moreover, if a sample is badly classified ($ p-> 0 $), the modulation factor is close to $ 1 $ and the cost function is little affected. The parameter $ \gamma $ therefore makes it possible to control the contribution of a sample in the final loss function according to its classification complexity.
In this study, we don't aim to study this cost function, we will not dwell on the choice of the hyperparameters $ \alpha $ and $ \gamma $. We use the optimal parameters obtained by  Johnson et~al. in \cite{johnson2019medicare}: $\alpha=0.25$ and $\gamma=3$. 

\subsubsection{MFE Loss}
The mean false error (MFE) cost function decomposes the mean squared error (MSE) into two components during optimization in order to give more weight to the minority class samples. 
The MSE is rewritten as a kind of weighted average of the errors of the two classes. In this way, all the classes participate equally in the final loss function. The final loss function is a sum of to means : \textbf{mean false positive error} (FPE) and \textbf{mean false negative error} (FNE).\\
$ MFE=FPE+ FNE$ and $MSFE=FPE^2+ FNE^2$

$$FPE=\frac{1}{N} \sum_{i=1}^N\sum_{n}\frac{1}{2}(d^{(i)}_{n}-y^{(i)}_{n})^2$$
$$FPE=\frac{1}{P} \sum_{i=1}^P\sum_{n}\frac{1}{2}(d^{(i)}_{n}-y^{(i)}_{n})^2$$
\textbf{N} is the number of samples in the negative class, $\textbf{P}$ number of samples in the positive class, $d^{(i)}$ the true label of sample $i$, $y^{(i)}$ the predicted label for sample $i$.

\subsection{Experimental Design }
The dataset has been separated into a training dataset (80 \%) and test (20 \%) dataset. To avoid any risk of data leakage, we split the dataset according to the healthcare provider Id. Thus, if a provider is in the training subset, his claims are only used in the training steps. The models are trained using a k-folds cross-validation ($ k = $ 10 in our case).
Each model is therefore trained and validated on 10 different sub-samples. 
The final performance is computed on the test set. Note that the test sets do not intervene at any time in the training steps. The final value of each performance metric is computed by taking the average of the ten measurements obtained during the ten iterations of the cross-validation. The final models are compared using the AUC (PRC) metric. We set the number of epochs to 50 for each model and used \textbf{early stopping} to avoid overfitting.

\section{Results and discussions}
\begin{figure}[h]
\begin{center}
\centerline{\includegraphics[width=0.6\textwidth]{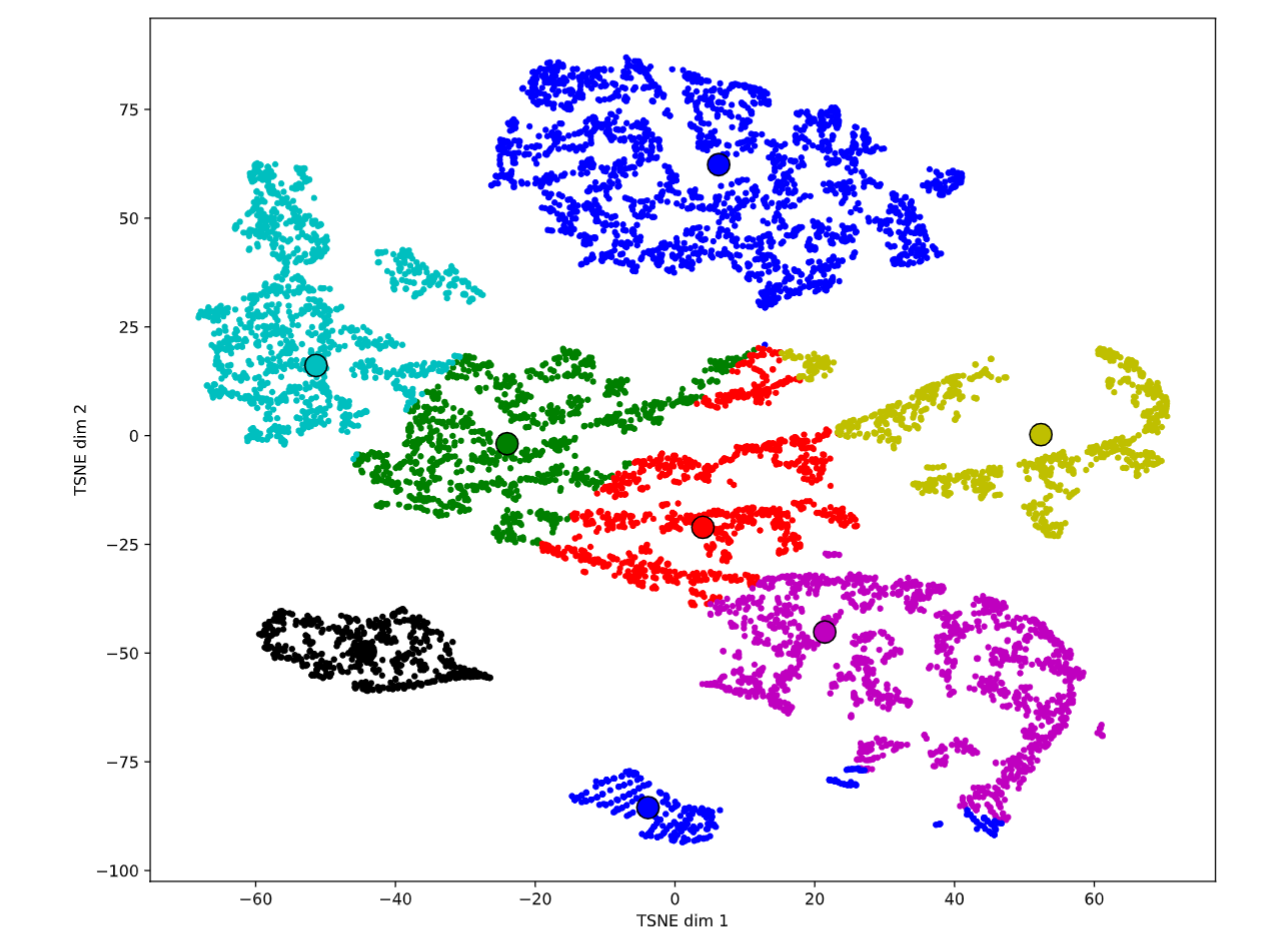}}
	\caption{Mean-shift  clustering on the autoencoder latent vector. This output of the autoencoder separates the providers into homogeneous groups.}
	\label{latent_projection}
\end{center}
\end{figure}

In this section we present the results of our classifiers on data sets from 2017 to 2019 collected on the CMS website \cite{CMSGov2018}. The classifiers performances metrics are listed in Table~\ref{results2}. 
Recall that we refer to the baseline neural network as \textbf{BNN} and our multiple inputs neural network as \textbf{MINN-AE}. \textbf{BNN weighted} stands for BNN with weighted loss, \textbf{BNN focal} with focal loss, \textbf{BNN mfe} with the mean false error loss and \textbf{BNN rus} the best BNN obtained by random under sampling.
Despite the class imbalance in the training data, MINN-AE outperform all other classifiers in terms of AUC (PRC). Our model's AUC (PRC) is 0.745 and that of the second best classifier is 0.616. Note that the \textbf{no skill} (random) classifier has an AUC (PCR) of 0.03. Using the ROC (AUC) as performance metric, the baseline neural network with mean false error function (BNN mfe) has the best performance (0.864) but it has a very low precision (0.445). This is due to the fact this model fails to capture the provider behavior and the context of the data. The advantage of MINN-AE is that the autoencoder separates the providers into homogeneous groups and creates contextual features. In fraud detection the context matters. For example two providers can provide very similar claims but due to their previous behaviors (context) one will be considered fraudulent and the other one genius.
State-of-the-art classifiers (logistic regression, random forest, Gradient boosting and XGBoost) perform worst than artificial neural network classifiers because they fails to capture complex structures in sequence datasets and large scale data \cite{chalapathy2019deep}. Deep learning models have excellent capabilities in learning expressive representations of complex data such as high-dimensional data, temporal data and spatial data \cite{pang2021deep}.
Our results suggest that using an LSTM autoencoder to embed the provider level features makes it easier for the neural network to separate fraudulent transaction from legitimate ones. The autoencoder acts like a dimensional reduction layer and also learns the provider behavior. 
The model is also robust toward the imbalance class due to the fact that the latent features extracted from the LSTM autoencoder have strong clustering power. 
The latent features  allows the model to group the providers into clusters (see Fig.~\ref{latent_projection}) and makes it easier to identify fraudulent behaviors. We can see on Fig.\ref{latent_projection} that the providers are clustered into homogeneous groups. This clustering was done using the \textbf{MeanShift} algorithm from \textbf{Scikit learn}. These results show that although it is important to take in account the provider level data in medicare fraud detection, it is also important to separate them from the features related to the claims when feeding an artificial neural network. 
 \begin{table}[htbp]
\caption{Experimental results of the proposed method and some state-of-the-art methods. Mean time refers to the execution time expressed in minutes.}
\label{results2}
\begin{center}
\resizebox{0.8\columnwidth}{!}{%
\begin{tabular}{|l|c|c|c|c|c|c|c|c|}
 \hline
{Models} & Precision & AUC(ROC) &  Gmean  & AUC (PRC)&Mean Time \\
 \hline
 Random classifier&- & 0 & 0.5 &0.03 & -\\
 Logistic regression&  0.438 &    0.827 &   0.826 &   0.629 &  0.51\\
Random forest& 0.599 &    0.807 &   0.796 &   0.658&  2.35 \\
Gradient boosting&0.713 &    0.715 &   0.666 &  0.617 &15.43  \\
XGBoost &0.707 &    0.713 &   0.663 &   0.613&0.53\\
BNN& 0.436 &    0.863 &   0.862 &   0.739 & 2.54\\
BNN  weighted \cite{johnson2019medicare}& 0.429 &    0.860 &   0.859 &   0.733 & 2.96\\
BNN   focal \cite{lin2017focal}&  0.432 &    0.861 &   0.861 &   0.737 & 2.50 \\
BNN   mfe \cite{wang2016training}& 0.445 &    0.864 &   0.863 &   0.741 & 6.03\\
BNN RUS & 0.534 &   0.720 &   0.668 & 0.512&1.20\\
\textbf{MINN-AE (Ours)} &\textbf{0.770}  &\textbf{0.794} &   \textbf{0.762} & \textbf{0.765}&10  \\
 \hline
\end{tabular}}
\end{center}

\end{table}

\section{\large\bfseries Conclusion}
Medicare fraud results in considerable losses for governments, insurance companies and taxpayers. 
Frauds or anomalies are very rare events and difficult to detect. In most of the cases, the collected data are streaming time series data and due to their intrinsic characteristics it is a challenging problem to detect frauds precisely in them. Performance of traditional algorithms in detecting frauds is sub-optimal on large  scale  data  since  they  fails  to  capture  complex  structures  in  the  data. Deep learning based models have shown excellent capabilities in learning expressive representations of complex data such as high-dimensional data, temporal data, spatial data and graph data.

In this study, we proposed a deep neural networks with multiple inputs called \textbf{MINN-AE} to detect medicare frauds. Our model has a LSTM based autoencoder  component that learns contextual features from the input data. Our results showed that this kind of architecture outperforms a classical multi-layer perceptron  models using a single input layer. The model is also robust toward the imbalance class due to the fact that the latent features extracted from the autoencoder have strong discriminating power. Future work will include employing the multiple inputs models with data sampling techniques or algorithm level techniques  to  combat  the  imbalanced  nature  of the data.
\bibliographystyle{unsrt} 
\bibliography{references}
\nocite{9680050}
\nocite{9674545}
\nocite{9643276}

\end{document}